\documentclass[preprint,12pt]{elsarticle}
\usepackage{amsmath,amsfonts}
\usepackage{array}
\usepackage{booktabs}
\usepackage{multirow}
\usepackage{placeins}
\usepackage{url}
\usepackage{graphicx}
\usepackage{float}
\usepackage{pdflscape}
\journal{Biomedical Signal Processing and Control}
\biboptions{sort&compress}
\begin{document}
\begin{frontmatter}

\title{Attribution-Gated Joint Estimation of Intention and Capability for Assist-as-Needed Upper-Limb Robots}

\author[aff1]{Pengcheng Li}
\author[aff1,aff2]{Shuxiang Guo\corref{cor1}}
\ead{guo.shuxiang@sustech.edu.cn}
\author[aff1]{Hanze Wang}
\author[aff1]{Jun Leng}
\author[aff3]{Bin Wang}
\author[aff3]{Mingchao Ding}
\cortext[cor1]{Corresponding author.}

\affiliation[aff1]{organization={Department of Electronic and Electrical Engineering, Southern University of Science and Technology},
            city={Shenzhen},
            country={China}}
\affiliation[aff2]{organization={Aerospace Center Hospital, School of Life Science, and the Key Laboratory of Convergence Medical Engineering System and Healthcare Technology, Ministry of Industry and Information Technology, Beijing Institute of Technology},
            city={Beijing},
            country={China}}
\affiliation[aff3]{organization={Department of Peripheral Vascular Intervention, Aerospace Center Hospital, School of Life Science, Beijing Institute of Technology},
            city={Beijing},
            postcode={100081},
            country={China}}

\begin{abstract}
Assist-as-needed (AAN) control must distinguish between changes in intended motion and changes in effective voluntary torque capability, even though both can produce similar kinematic and interaction responses. A dual-measurand multimodal MHE--MPC architecture is developed as the base framework, in which the intended trajectory and the task-dependent effective voluntary torque capability are estimated online as distinct quantities, and both estimates are used to adapt robotic assistance. Building on this architecture, the ambiguity between the two update channels is formulated as a continuous credit-assignment problem and resolved by an attribution gate. A trajectory-change indicator is obtained from persistent corrective interaction, whereas a capability-change indicator is derived from the residual between normalized muscle activation and dynamics-reconstructed voluntary torque. 
Nine healthy participants completed factorial target-height–load trials and a pre–post fatigue cup-transport task. Without target-height input, the gated method generated correctly ordered Low–Middle–High references under every load: target changes were preferentially assigned to trajectory adaptation, whereas added load shifted adaptation toward the capability channel. It achieved the lowest RMSE in all nine conditions (0.017 rad versus 0.041 rad for ungated dual estimation and 0.127 rad for fixed-reference error-based AAN). Post-fatigue capability estimates decreased and assistance increased at all heights, supporting functional feasibility.
\end{abstract}

\begin{keyword}
Attribution gating, Assist-as-needed control, Motion intention estimation, 
Voluntary torque capability, Model predictive control.
\end{keyword}

\end{frontmatter}

\section{Introduction}
\label{sec:introduction}
Upper-limb motor impairment creates demand for intensive and task-specific rehabilitation training~\cite{Tian2025,Brown2024}. Repeatable movement practice can be delivered and human--robot interaction can be quantified by rehabilitation robots during training~\cite{Brown2024}. Robot-mediated rehabilitation can be further extended by wearable platforms toward functional tasks and less structured environments~\cite{Li2025,PLi2025,Feng2025,Proietti2024}.

Task completion can be facilitated by assist-as-needed (AAN) control while the user's voluntary contribution is preserved. This balance is particularly difficult in wearable systems, which operate in continuous physical interaction with users whose movement goals, effort, fatigue, and physical limitations may vary across repetitions~\cite{ZYang2025, LiHe2025JBHI}. A reference trajectory that is inconsistent with the user's intended movement can produce tracking error and interaction torque, but similar responses can also arise when voluntary capability is insufficient for the required motion. Active participation may be suppressed by excessive assistance, whereas task completion may be compromised or interaction load may be increased by insufficient assistance. Effective AAN control must therefore address both trajectory compatibility and assistance demand.

Trajectory compatibility has been addressed by one line of research through human-intention estimation and reference adaptation. sEMG, kinematics, and interaction torque are combined with model-based or data-driven inference by multimodal methods~\cite{Zhou2025, Wu2025, Han2025, Franceschi2025, Yang2024, Xiong2024}; the reference is further adapted by iterative learning and optimization~\cite{Ozen2022, Han2025, Franceschi2025}. Motion intention and arm stiffness have also been estimated simultaneously for variable-impedance interaction~\cite{Wu2025}. However, stiffness is an interaction property rather than the voluntary torque effectively available for the current task. To the best of the authors' knowledge, intended trajectory and task-dependent effective voluntary torque capability have not been jointly estimated as distinct online measurands for AAN. Moreover, trajectory-adaptation signals depend on whether the user can execute the intended movement.

Assistance is modulated by error-based AAN schemes from kinematic deviation~\cite{Zhang2025TwoLayer, Garcia2025Velocity, ZYang2023Task, Xiong2022Elbow}, whereas voluntary torque, stiffness, or strength are estimated by capability-centered methods to compensate for an inferred deficit~\cite{Sharifi2025RL, Shah2025Shoulder, Zarrin2024VariableAdmittance, Sedighi2024EMGDL, Cao2024AdaptivePosition}. Physiological effort information is provided by direct sEMG-based assistance, although signal variability and session-specific calibration remain practical considerations~\cite{Sedighi2024EMGDL, Cao2024AdaptivePosition}. How much assistance should be provided is determined by these approaches, but tracking and effort deviations may also reflect a purposeful change in the desired trajectory.

Trajectory adaptation and capability-based assistance consequently cannot be treated as unrelated estimation problems. The task demand is modified by a change in intention, while the effective voluntary contribution available in the current task context is modified by fatigue, activation--torque inconsistency, and unmodeled external demand. These effects can occur during the same movement and appear in the same motion, interaction, and effort observations. If the two states are updated from overlapping residuals without explicit credit assignment, capability- or demand-related signals can be absorbed into trajectory adaptation, while intention-related signals can be absorbed into assistance.

To address this problem, a dual-estimation architecture is established and then extended by automatic attribution. The contributions are as follows:

\noindent (1) An ungated dual-measurand MHE--MPC architecture is developed to estimate intended-trajectory deformation and task-dependent effective voluntary torque capability online, and both estimates are used for AAN.

\noindent (2) An attribution gate is introduced to compensate cross-sensitivity by allocating shared observations through a gating variable $\alpha \in [0,1]$ and its complement $1-\alpha$, thereby limiting leakage between intention and capability estimates.

\noindent (3) The two layers are evaluated separately: dual estimation is tested by comparison with error-based AAN, whereas credit assignment is tested by gated--ungated comparison.

\section{Measurement System and Estimated Quantities}
\subsection{Robot and Sensor Platform}
\label{sec:robot}

\begin{figure}[H]
    \centering
    \includegraphics[width=0.82\textwidth]{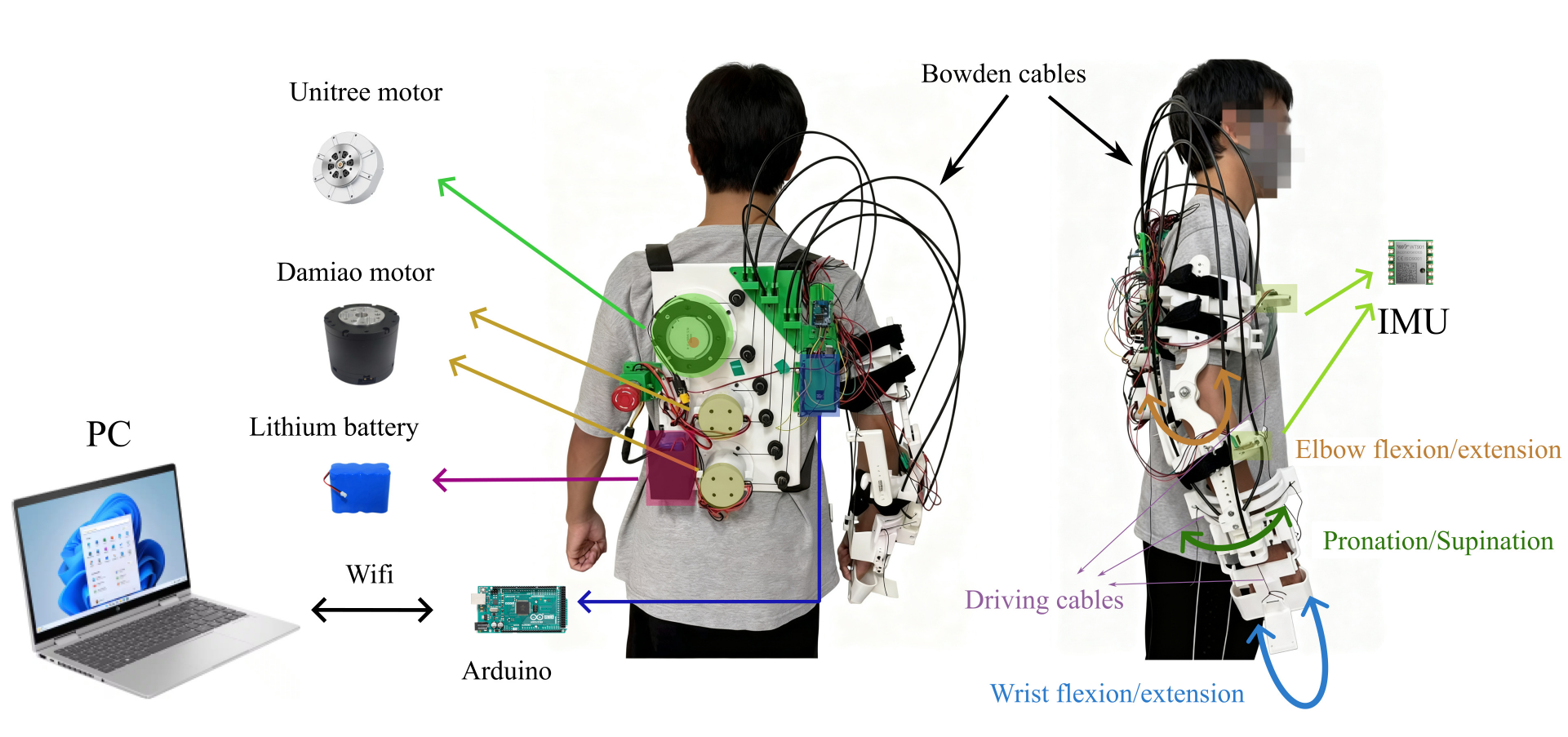}
\caption{Wearable upper-limb robot and control hardware. Cable-driven modules
assist elbow and wrist motion, IMUs measure joint angles, and ROS~2 communicates
with the embedded motor drivers.}
    \label{fig:Hardware}
\end{figure}

A lightweight, portable upper-limb wearable robot was developed as the experimental platform for elbow and wrist assistance, as depicted in Fig.~\ref{fig:Hardware}. The device employs a cable-driven mechanism to deliver assistive torque to the elbow and wrist. A Unitree motor (GO-M8010-6, YuShu Technology Co., Hangzhou, China) actuates the elbow module, while two Damiao motors (DM-J4310-2EC, Damiao Technology Co., Shenzhen, China) actuate wrist flexion/extension and pronation/supination. Inertial measurement units (IMUs) (JY-901B, WitMotion, Shenzhen, China) mounted on the robot provide estimates of joint angles. An Arduino Mega 2560 microcontroller (Arduino, Ivrea, Italy) interfaces with the motor drivers and IMUs, executing the embedded torque control loop at 500~Hz. A laptop runs the upper-level dual-estimation MPC on ROS~2 at 50~Hz, communicating with the Arduino over a local Wi-Fi link. In this setup, the effective one-way communication latency was approximately 20~ms (about one upper-level sampling interval). Because the high-bandwidth torque loop is closed on the Arduino and the upper-level assistance/trajectory-deformation updates are rate-limited and dominated by slow task dynamics (typically 1--10~Hz), this latency mainly acts as a one-step hold on slowly varying commands. A lithium-ion battery powers the system, enabling untethered operation for extended periods.

To improve wearability, the actuators are placed away from the distal joints and the assistive torques are transmitted to the elbow and wrist through cables. The robot has a total mass below 5~kg and a wearable mass of approximately 0.7~kg. Adjustable straps support rapid fitting across participants.

\subsection{Sensor Signals and Voluntary Torque Reconstruction}

In the joint-level representation of the cable-driven robot, $\boldsymbol{\tau}_{robot}$ denotes the effective joint-side torque returned by the embedded low-level actuation loop. Unmodeled cable compliance, friction, intermittent slack, linearization error, and sensing error are included in the lumped disturbance term $\boldsymbol{d}(t)$ used in Section~\ref{sec:stability_feasibility}. The robot dynamics in joint space are modeled as
\begin{equation}
\begin{aligned}
    \mathbf{M}_R \ddot{\boldsymbol{\theta}} + \mathbf{C}_R(\boldsymbol{\theta}, \dot{\boldsymbol{\theta}}) \dot{\boldsymbol{\theta}} + \mathbf{G}_R(\boldsymbol{\theta}) 
    = \boldsymbol{\tau}_{robot} - \boldsymbol{\tau}_{HRI},
\end{aligned}
\label{eq:robot_dyn}
\end{equation}
where $\boldsymbol{\theta}$ is the joint angle vector, and $\mathbf{M}_R$, $\mathbf{C}_R$, and $\mathbf{G}_R$ are the robot's inertia matrix, Coriolis/centrifugal matrix, and gravity vector, respectively. $\boldsymbol{\tau}_{robot}\in\mathbb{R}^n$ is the commanded motor torque, and $\boldsymbol{\tau}_{HRI}\in\mathbb{R}^n$ denotes the generalized human--robot interaction torque in joint space. The signs of $\boldsymbol{\tau}_{HRI}$ in Eqs.~(\ref{eq:robot_dyn}) and~(\ref{eq:human_dyn}) follow the action--reaction convention at the cuff interface.

The dynamics of the human limb are expressed as:
\begin{equation}
\begin{aligned}
    \mathbf{M}_{H} \ddot{\boldsymbol{\theta}} + \mathbf{C}_{H}(\boldsymbol{\theta}, \dot{\boldsymbol{\theta}}) \dot{\boldsymbol{\theta}} + \mathbf{G}_{H}(\boldsymbol{\theta}) =  \boldsymbol{\tau}_{vol} + \boldsymbol{\tau}_{HRI},
\end{aligned}
\label{eq:human_dyn}
\end{equation}
where $\boldsymbol{\tau}_{vol}$ represents the net voluntary torque generated by the user, and $\mathbf{M}_{H}$, $\mathbf{C}_{H}$, and $\mathbf{G}_{H}$ are the human limb's inertia matrix, Coriolis/centrifugal matrix, and gravity vector, respectively. Under the chosen sign convention, the measured interaction torque is decomposed as:
\begin{equation}
    \boldsymbol{\tau}_{HRI} = \boldsymbol{\tau}_{Assi} - \boldsymbol{\tau}_{corr},
    \label{equ:ori_HRI}
\end{equation}
where $\boldsymbol{\tau}_{Assi}$ is the task-aligned assistive torque from the robot, and $\boldsymbol{\tau}_{corr}$ denotes the corrective interaction torque relative to the current reference.

\subsection{Motion Intention and Capability Variables}

The user's intended motion is represented as a smooth deformation of a nominal joint-space task trajectory:
\begin{equation}
    \boldsymbol{\theta}_{int}(s)
    =
    \boldsymbol{\theta}_{nom}(s)
    +
    \boldsymbol{\Phi}(s)\mathbf{w},
    \label{eq:intent_traj_param}
\end{equation}
where $s\in[0,1]$ is the task phase, $\boldsymbol{\theta}_{nom}(s)$ is the nominal joint trajectory, $\boldsymbol{\Phi}(s)$ is a fixed smooth basis, and $\mathbf{w}$ is the trajectory-deformation parameter vector estimated online. The vector $\mathbf{w}$ encodes the amplitude and direction of spatial path deformation, such as raising or lowering the endpoint relative to the nominal reach.

The effective voluntary capability is represented by a vector-valued, control-oriented torque envelope $\boldsymbol{\tau}_{max}$. Each component corresponds to one controlled joint direction. This state represents the voluntary torque effectively available under the current task and interaction conditions. It may therefore reflect fatigue, activation--torque consistency, and external demand not represented separately in the prediction model. Given a latent task-demand torque $\boldsymbol{\tau}_{desire}$, the available voluntary contribution is approximated by an element-wise saturation abstraction:
\begin{equation}
\begin{aligned}
\boldsymbol{\tau}_{vol} &= \mathcal{S}(\boldsymbol{\tau}_{desire};\,\boldsymbol{\tau}_{max}),\\
[\mathcal{S}(x;\tau_{max})]_i &= \mathrm{sign}(x_i)\,\min\{ |x_i|,\,\tau_{max,i} \}.
\end{aligned}
\label{eq:saturation}
\end{equation}
The net voluntary contribution is decomposed as
\begin{equation}
    \boldsymbol{\tau}_{vol} = \boldsymbol{\tau}_{task} + \boldsymbol{\tau}_{corr},
    \label{eq:vol_decompose}
\end{equation}
where $\boldsymbol{\tau}_{task}$ supports task progression and $\boldsymbol{\tau}_{corr}$ represents corrective interaction relative to the current reference. These representations provide the state coordinates used by the credit-assignment controller in Section~III.

\section{Joint Estimation of Intention and Capability}
\subsection{Problem Definition}
At control step $k$, the two latent measurands are stacked in
\begin{equation}
\mathbf z_k=
\begin{bmatrix}
\mathbf w_k^T & \boldsymbol{\tau}_{max,k}^T
\end{bmatrix}^{T},
\label{eq:latent_measurands}
\end{equation}
where $\mathbf w_k$ parameterizes intended-trajectory deformation and $\boldsymbol{\tau}_{max,k}$ denotes the vector-valued, task-dependent effective voluntary capability. Each component of $\boldsymbol{\tau}_{max,k}$ corresponds to the magnitude of voluntary torque that the user can reliably produce in one controlled joint direction under the current task context. Measured motion, interaction, and muscle activity provide the multimodal observation
\begin{equation}
    \mathbf{y}_k=
    \left\{
    \boldsymbol{\theta}_k,\dot{\boldsymbol{\theta}}_k,
    \hat{\boldsymbol{\tau}}_{corr,k},
    \boldsymbol{\tau}^{EMG}_{vol,k}
    \right\}.
    \label{eq:credit_assignment_observations}
\end{equation}
Here $\boldsymbol{\tau}^{EMG}_{vol,k}$ is the voluntary torque estimated from sEMG; it is decomposed in the calibration stage (Section~\ref{sec:participant_calibration}) into a normalized activation $\bar a_{p,k}$ and a dynamics-reconstructed mechanical voluntary torque $\tau^{mech}_{p,k}$, which are the quantities used in subsequent estimation steps.

The measurement problem is to infer $\mathbf z_k$ online from a recent window of $\mathbf y_k$. The effects of its two components overlap: deviation from the nominal path may reflect either a different movement goal or insufficient capacity to realize that goal. Defining the errors relative to the nominal and intended paths gives
\begin{equation}
    \boldsymbol{e}_{nom,k}
    =
    \boldsymbol{e}_{int,k}
    +\boldsymbol{\Phi}(s_k)\mathbf{w}_k.
    \label{eq:nominal_error_decomposition}
\end{equation}
Thus, nominal tracking error alone cannot determine whether the reference or the assistance should change. Misclassifying an intentional trajectory change as a loss of capability produces unnecessary assistance, whereas misclassifying a capability loss as a new intention may deform the reference and withhold needed support. The control problem is therefore to identify trajectory intention and voluntary capability simultaneously.

\subsection{System Overview}

\begin{figure}[H]
    \centering
    % Canonical editable SVG: Figure/Figure2_Method_clear.svg
    % The matching PDF below is used because XeLaTeX does not embed SVG directly.
    \includegraphics[width=0.92\textwidth]{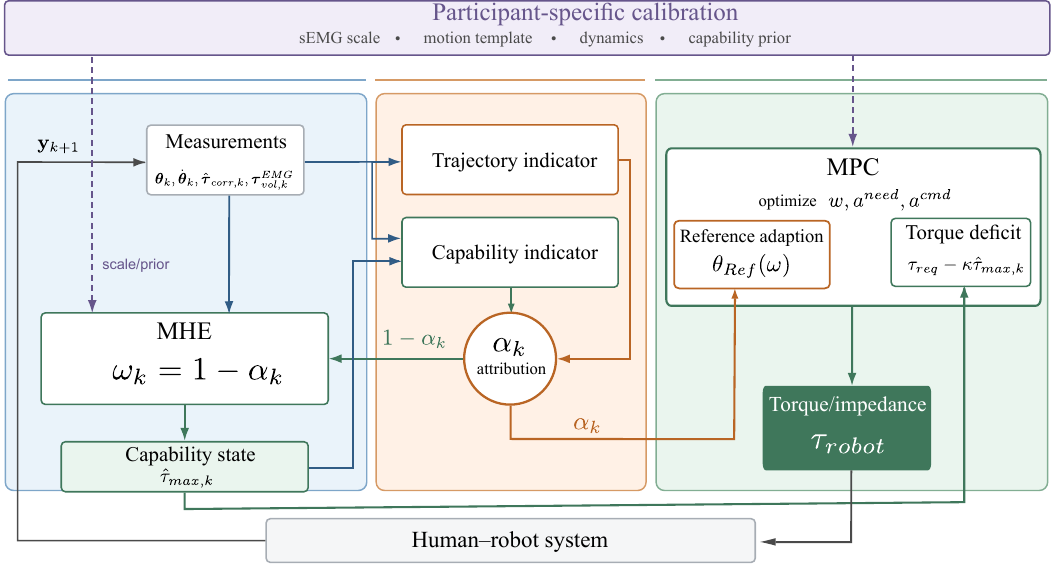}
    \caption{Two-layer joint estimation and assistive-control method. Base MHE and MPC estimate capability and intended trajectory, respectively; both estimates determine assistance. The added gate is updated first: $1-\alpha$ weights capability data and $\alpha$ weights trajectory correction.}
    \label{fig:control_architecture}
\end{figure}

The method comprises three layers (Fig.~\ref{fig:control_architecture}). In the base architecture, capability is estimated by MHE and intended-trajectory deformation by MPC, and both estimates determine assistance. The extension uses $\boldsymbol{\alpha}_k$ to gate their updates. Participant-specific quantities are obtained from the calibration in Section~\ref{sec:participant_calibration}; synchronized kinematics, sEMG, and robot torque are then used to reconstruct the participant's mechanical contribution.

The central, joint-direction-specific attribution state $\boldsymbol{\alpha}_k$ compensates cross-sensitivity between the two measurement channels. Persistent corrective interaction supports a change in movement intention and therefore updates the reference trajectory, whereas disagreement between muscle activation and model-implied voluntary torque supports a change in capability. Importantly, $\boldsymbol{\alpha}_k$ is a supervisory state, not an MPC decision variable. At each cycle it is computed before the two optimizations: $1-\boldsymbol{\alpha}_k$ gates the capability indicator in MHE, whereas $\boldsymbol{\alpha}_k$ gates the trajectory indicator in MPC.

Moving-horizon estimation (MHE) uses a recent window of measurements to update the effective-capability estimate under reliability, physiological, and rate constraints~\cite{Rao2003MHE}. The accepted capability is then supplied to model predictive control (MPC), which predicts future motion and jointly optimizes reference deformation and the required assistance subject to joint, torque, and slew-rate limits~\cite{Mayne2000MPC}. Thus, MHE estimates what the participant can effectively contribute under the current task context from the past observation horizon, while the attribution-conditioned MPC estimates intended trajectory over the predictive horizon and jointly determines the remaining assistance demand.

\subsection{Separating Intention and Capability Effects}
\subsubsection{Trajectory Update From Corrective Torque}
The joint-side interaction torque is reconstructed from the identified robot dynamics,
\begin{equation}
    \hat{\boldsymbol{\tau}}_{HRI}
    =
    \boldsymbol{\tau}_{robot}
    -
    \big(
    \mathbf{M}_{R}\ddot{\boldsymbol{\theta}}
    + \mathbf{C}_{R}\dot{\boldsymbol{\theta}}
    + \mathbf{G}_{R}
    \big),
    \label{eq:hri_est}
\end{equation}
and the component opposing the current assistance is
\begin{equation}
    \hat{\boldsymbol{\tau}}_{corr}
    = \boldsymbol{\tau}_{Assi}-\hat{\boldsymbol{\tau}}_{HRI}.
    \label{equ:tau_corr}
\end{equation}
Using the human-limb dynamics in Eq.~(\ref{eq:human_dyn}), the mechanical voluntary torque is reconstructed as
\begin{equation}
\begin{aligned}
\hat{\boldsymbol{\tau}}^{mech}_{vol,k}
= {}&\mathbf{M}_{H}(\boldsymbol{\theta}_k)\ddot{\boldsymbol{\theta}}_k
+\mathbf{C}_{H}(\boldsymbol{\theta}_k,\dot{\boldsymbol{\theta}}_k)
\dot{\boldsymbol{\theta}}_k\\
&+\mathbf{G}_{H}(\boldsymbol{\theta}_k)
-\hat{\boldsymbol{\tau}}_{HRI,k}.
\end{aligned}
\label{eq:mechanical_voluntary_torque}
\end{equation}
For participant $p$, $\tau^{mech}_{p,k}$ denotes the corresponding joint-direction component of this reconstructed torque.
To suppress brief impacts and model error, the corrective torque is passed through a normalized, leaky observer. Let $\boldsymbol{\tau}_{norm,k}$ be the calibration-derived maximum voluntary torque for the current joint, used to nondimensionalize the corrective interaction. The observer is constructed as
\begin{equation}
\begin{aligned}
\bar{\boldsymbol{\tau}}_{corr,k}
&=
\mathrm{sat}_{[-1,1]}
\left(
\frac{\mathcal{D}_{\epsilon}(\hat{\boldsymbol{\tau}}_{corr,k})}
{\boldsymbol{\tau}_{norm,k}+\epsilon_0}
\right),\\
\dot{\boldsymbol{\chi}}_{k}
&=(1-\lambda_{\chi}\Delta t)\dot{\boldsymbol{\chi}}_{k-1}
+\Delta t\,\mathbf{B}_{\chi}\bar{\boldsymbol{\tau}}_{corr,k},\\
\boldsymbol{\chi}_{k}
&=\Pi_{\mathcal{X}}
\left(
\boldsymbol{\chi}_{k-1}+\Delta t\dot{\boldsymbol{\chi}}_{k}
\right),
\end{aligned}
\label{eq:corr_observer}
\end{equation}
where $\mathcal D_{\epsilon}$ is a dead-zone operator, $\mathbf{B}_{\chi} = \mathrm{diag}(b_{\chi,1},\ldots,b_{\chi,n})$ with $b_{\chi,i}>0$ is a positive diagonal gain matrix that controls the integration rate for each joint direction, $\epsilon_0>0$ only guards the normalization against near-zero $\boldsymbol\tau_{norm,k}$, and $\Pi_{\mathcal X}$ enforces the admissible deformation range. This construction follows standard robust leaky-integrator observer design: the dead zone rejects sub-threshold noise, and the exponential forgetting in $\dot{\boldsymbol\chi}_k$ bounds the observer state under persistent excitation. The bounded state $\boldsymbol{\chi}_k$ therefore indicates a persistent, rather than impulsive, need to change the reference path.

\subsubsection{Capability Estimation}
\label{sec:capability_estimation}
The participant-specific activation scale and initial effective-capability prior are obtained from calibration (Section~\ref{sec:participant_calibration}). Let $\bar a_{p,k}$ denote the resulting normalized task activation and $\tau^{mech}_{p,k}$ the corresponding voluntary torque reconstructed using the participant-specific dynamics. The calibration-derived effective capability is used only as the prior for online estimation. The linear scaling $\bar a_{p,j}\tau$ in the MHE residual is a local approximation of the voluntary torque--activation relationship validated during calibration. 

Online, the effective-capability estimate is updated over a moving window $\mathcal W_k$. Each sample $j$ contributes to the capability estimate only to the extent that it is attributed to a capability-related change, with the weight given directly by the complement of the trajectory-attribution gate:
\begin{equation}
\omega_j = 1-\alpha_j .
\label{eq:capability_observation_gate}
\end{equation}
For one joint-direction component, the moving-horizon estimate is then
\begin{equation}
\begin{aligned}
\tilde\tau_{max,k}
=\arg\min_{\tau\in\mathcal T}\;&
\lambda_a(\tau-\hat\tau_{max,k-1})^2\\
&+\sum_{j\in\mathcal W_k}\omega_j
\mathcal{H}_{\delta_c}
\left(\tau^{mech}_{p,j}-\bar a_{p,j}\tau\right),\\
\hat\tau_{max,k}
&=\Pi_{\mathcal T,\dot\tau}
\left(\tilde\tau_{max,k}\right).
\end{aligned}
\label{eq:capability_mhe}
\end{equation}
Here $\mathcal T = [0, \tau_{max}^{phys}]$ is the physiologically plausible torque range for the current joint direction, and $\Pi_{\mathcal T,\dot\tau}$ additionally enforces a maximum rate of change $\dot{\tau}_{max}^{max}$ between consecutive estimates. The arrival term preserves continuity in the moving-horizon sense of~\cite{Rao2003MHE}, and the Huber loss $\mathcal H_{\delta_c}$ is the standard M-estimator that interpolates between quadratic and linear penalties to limit the influence of EMG and acceleration outliers.

\subsection{Joint Estimation and Assistive Control}
% \subsubsection{MHE--MPC Coordination}
% The two measurands are updated over coupled temporal horizons. MHE infers capability from physiological--dynamic consistency over the past observation horizon, while attribution-conditioned MPC estimates the intended-trajectory deformation and predicts motion over the future horizon. Separating their indicator channels prevents the tracking objective from altering the capability state merely to reduce control error: capability changes must be supported by EMG--dynamics data. Within this integrated predictive measurement-and-control layer, $\mathbf w$ represents the inferred movement intention, the nonnegative deficit $\mathbf a^{need}$ represents demand beyond voluntary capability, and $\mathbf a^{cmd}$ represents the constrained assistance required to address that deficit.

\subsubsection{Attribution Weight Update}
The trajectory indicator $E^{traj}_k$ is derived from persistent corrective interaction, whereas the capability indicator $E^{cap}_k$ is derived from the dynamics--EMG residual evaluated at the previous accepted capability. For one joint direction, the two nonnegative indicators are
\begin{equation}
\begin{aligned}
E^{traj}_k&=|\chi_k|,\\
r^{cap}_j(\tau)&=\tau^{mech}_{p,j}-\bar a_{p,j}\tau,\\
\bar r^{cap}_j(\tau)&=
\frac{r^{cap}_j(\tau)}{\tau_{scale,p}+\epsilon_\tau},\\
E^{cap}_k&=
\frac{\displaystyle\sum_{j\in\mathcal W_k}\omega_j
\mathcal H_{\delta_c}\!\left(
\bar r^{cap}_j(\hat\tau_{max,k-1})\right)}
{\displaystyle\sum_{j\in\mathcal W_k}\omega_j+\epsilon_E}.
\end{aligned}
\label{eq:attribution_evidence}
\end{equation}
Here $\tau_{scale,p}>0$ is the calibration-derived torque scale (see Section~\ref{sec:participant_calibration}), which together with the normalized corrective-torque observer in Eq.~(\ref{eq:corr_observer}) makes the two indicator channels dimensionless. The weights $\omega_j$ in the summation use the attribution value $\alpha_j$ stored at the time sample $j$ was acquired, not the current $\alpha_k$, thus decoupling the two updates and preventing an algebraic loop. The small constants $\epsilon_\tau,\epsilon_E>0$ appearing above only guard against division by near-zero denominators and do not otherwise affect the estimator. The ratio of the two indicators defines the attribution target:
\begin{equation}
\begin{aligned}
\alpha^{tar}_k&=
\begin{cases}
\dfrac{E^{traj}_k+0.5E_0}
{E^{traj}_k+E^{cap}_k+E_0},
&E^{traj}_k+E^{cap}_k\ge\epsilon_{\alpha},\\
0.5,&\text{otherwise},
\end{cases}\\
\alpha_k&=\Pi_{[0,1],\dot{\alpha}}
\left(\alpha_{k-1}
+\lambda_{\alpha}(\alpha^{tar}_k-\alpha_{k-1})\right).
\end{aligned}
\label{eq:attribution_gate_update}
\end{equation}
In this update, $E_0>0$ is a small baseline-evidence constant that acts as a pseudo-count prior to prevent weak observations from driving the gate to an extreme, $\epsilon_{\alpha}>0$ is a threshold that forces neutral attribution when neither indicator provides sufficient evidence, and $\dot{\alpha}^{max}$ is the maximum rate of change enforced by the projection $\Pi_{[0,1],\dot{\alpha}}$. 

In addition to gating the MHE observations through $1-\boldsymbol{\alpha}_k$, attribution complementarily shapes the trajectory-deformation residual in the MPC:
\begin{equation}
\begin{aligned}
\mathbf D_{\alpha,k}
&=\operatorname{diag}(\boldsymbol{\alpha}_k),\\
\mathbf r^{traj}_{i|k}
&=\mathbf D_{\alpha,k}\boldsymbol{\chi}_k
-\boldsymbol{\Phi}(s_{i|k})\mathbf w_{i|k},\\
J_{\mathrm{traj}}
&=\sum_{i=0}^{N-1}
\left\|\mathbf r^{traj}_{i|k}\right\|_{\mathbf Q_g}^{2}.
\end{aligned}
\label{eq:trajectory_gate_cost}
\end{equation}

\subsubsection{Reference Update and Assistance Calculation}
The participant-specific nominal trajectory is obtained from calibration. It is expressed by a smooth progression $\beta_p(s)$ from the comfortable start posture $\boldsymbol{\theta}_{0,p}$ to the middle-target endpoint $\boldsymbol{\theta}_{h_1,p}$:
\begin{equation}
\begin{aligned}
    \boldsymbol{\theta}_{nom,p}(s)
    &=\boldsymbol{\theta}_{0,p}
    +\beta_p(s)\Delta\boldsymbol{\theta}_p,\\
    \Delta\boldsymbol{\theta}_p
    &=\boldsymbol{\theta}_{h_1,p}-\boldsymbol{\theta}_{0,p},
    \qquad \beta_p(0)=0,\quad\beta_p(1)=1.
\end{aligned}
\label{eq:nominal_traj}
\end{equation}
All target conditions share this nominal trajectory; target height is shown to the participant but is not supplied to the controller. The intended path must therefore be inferred online through
\begin{equation}
    \boldsymbol{\theta}_{Ref,\,i|k}
    =
    \boldsymbol{\theta}_{nom,p}(s_{i|k})
    +
    \gamma_{traj}\boldsymbol{\Phi}(s_{i|k})\mathbf{w}_{i|k},
    \label{eq:ref_adaptation}
\end{equation}
where $\gamma_{traj}$ bounds the applied deformation. The corresponding task torque is predicted from the coupled dynamics:
\begin{equation}
\label{eq:tau_req_pred}
    \boldsymbol{\tau}_{req,\,i|k} = \mathbf{M}_0\,\ddot{\boldsymbol{\theta}}_{Ref,\,i|k} + \mathbf{C}_0\,\dot{\boldsymbol{\theta}}_{Ref,\,i|k} + \mathbf{G}_0,
\end{equation}
The nonnegative deficit $\mathbf a^{need}_{i|k}$ and rate-limited command $\mathbf a^{cmd}_{i|k}$ are retained separately so that assistance can change smoothly without discarding the estimated demand.

\subsubsection{MPC Cost Function and Constraints}
MPC is implemented as a receding-horizon quadratic program. Starting from the current measured joint state $\mathbf x_k=[\boldsymbol{\theta}_k^T, \dot{\boldsymbol{\theta}}_k^T]^T$, the discretized coupled dynamics predict
\begin{equation}
\begin{aligned}
\mathbf x_{0|k}&=\mathbf x_k,\\
\mathbf x_{i+1|k}
&=\mathbf f_d\!\bigl(
\mathbf x_{i|k},
\mathbf a^{cmd}_{i|k},
\\[-1pt]
&\qquad
\boldsymbol{\theta}_{Ref,i|k},
\hat{\boldsymbol{\tau}}_{max,k}
\bigr),
\quad i=0,\ldots,N-1.
\end{aligned}
\label{eq:mpc_state_prediction}
\end{equation}
The capability estimate is constant only within this short prediction horizon; it may be updated by MHE at the next measured cycle. The prediction therefore evaluates how a candidate reference deformation and assistance sequence would affect future motion, rather than assuming that the current tracking error remains unchanged.

The finite-horizon cost balances tracking, assistance, smooth adaptation, and attribution consistency:
\begin{equation}
\label{eq:mpc_cost}
\begin{aligned}
J_k^{MPC} = &\sum_{i=0}^{N-1} \Big(
\|\boldsymbol{e}_{R,\,i|k}\|_{\mathbf{Q}_e}^2
+\rho\,\|\mathbf{a}^{need}_{i|k}\|^2 \\
&\qquad + q_a\|\mathbf{a}^{cmd}_{i|k}
-\mathbf{a}^{need}_{i|k}\|^2
+\eta_{w}\,\|\Delta\mathbf{w}_{i|k}\|^2 \Big) \\
&+ \|\boldsymbol{e}_{R,\,N|k}\|_{\mathbf{P}}^2
+J_{\mathrm{traj}}.
\end{aligned}
\end{equation}
where $\boldsymbol e_R$ is the reference-tracking error. Penalizing assistance preserves the AAN objective, while the remaining terms prevent abrupt changes in reference or torque.

Joint, deformation, and torque bounds are imposed as
\begin{equation}
\label{eq:mpc_constraints}
\begin{aligned}
& \boldsymbol{\theta}_{Ref,\,i|k} \in \mathcal{C}, \\
& \mathbf{a}^{need}_{i|k} \ge \mathbf{0},\quad
\mathbf{a}^{need}_{i|k}\ge
\boldsymbol{\tau}_{req,\,i|k}
-\kappa\hat{\boldsymbol{\tau}}_{max,k}, \\
& \mathbf{0}\le\mathbf{a}^{cmd}_{i|k}\le\tau^{max}_{Assi},\\
&
\|\mathbf{a}^{cmd}_{i|k}-\mathbf{a}^{cmd}_{i-1|k}\|
\le \dot{\tau}_{Assi}^{max}\,\Delta t, \\
& \|\boldsymbol{\Phi}(s_{i|k})\mathbf{w}_{i|k}\|\le \theta_{def}^{max}, \\
& \|\mathbf{w}_{i|k}-\mathbf{w}_{i-1|k}\|
\le \dot{w}^{max}\,\Delta t,\\
&\|\ddot{\boldsymbol\theta}_{Ref,i|k}\|
\le \ddot\theta_{Ref}^{max}.
\end{aligned}
\end{equation}
where $\mathcal C$ is the safe joint set and $\kappa\in[0,1]$ is the fraction of estimated voluntary capability used as the assistance-onset threshold. The acceleration bound on $\ddot{\boldsymbol\theta}_{Ref}$ is enforced through the smoothness of the deformation basis $\boldsymbol{\Phi}(s)$ and the slew-rate limit on $\mathbf{w}$. Since $\mathbf a^{need}$ is nonnegative and penalized, the optimizer selects the smallest feasible deficit. Assistance is therefore triggered by predicted task demand exceeding $\kappa\hat{\boldsymbol{\tau}}_{max,k}$, rather than by raw EMG or tracking error. After solving Eqs.~(\ref{eq:mpc_cost})--(\ref{eq:mpc_constraints}), only $\mathbf a^{cmd\star}_{0|k}$ is sent to the lower-level controller and the first deformation value updates the reference. The remaining predicted sequence is used only as a warm start. At $k+1$, new measurements replace the prediction origin and the entire MHE--MPC sequence is repeated. This feedback update allows model mismatch and voluntary behavior changes to be corrected continually while retaining explicit joint and torque constraints.

\subsubsection{Torque--Impedance Assistance}
The optimal first-step command from the MPC is taken as the assistive torque, i.e., $\boldsymbol\tau_{Assi} = \mathbf a^{cmd\star}_{0|k}$. This assistance is combined with compliant impedance feedback:
\begin{equation}
\label{eq:robot_torque}
\boldsymbol{\tau}_{robot}
=-\mathbf K_R\boldsymbol e_R-\mathbf B_R\boldsymbol e_v
+\boldsymbol{\tau}_{Assi},
\qquad \mathbf K_R\succ0,\ \mathbf B_R\succ0 .
\end{equation}

\subsection{Tracking Stability and Command Boundedness}
\label{sec:stability_feasibility}
Let $\mathbf M=\mathbf M_R+\mathbf M_H$, $\mathbf C=\mathbf C_R+\mathbf C_H$, and $\mathbf G=\mathbf G_R+\mathbf G_H$. On the compact rehabilitation workspace, assume that $\mathbf M$ is uniformly positive definite and bounded and that $\dot{\mathbf M}-2\mathbf C$ is skew-symmetric. The projections and MPC constraints bound $\boldsymbol\alpha$, $\hat{\boldsymbol\tau}_{max}$, $\boldsymbol\theta_{Ref}$, its first two derivatives, and $\boldsymbol\tau_{Assi}$. Voluntary torque $\boldsymbol\tau_{vol}$ is saturated by the bounded envelope $\hat{\boldsymbol\tau}_{max,k}$ (whose magnitude is physically limited by $\tau_{max}^{phys}$) and is therefore also bounded. Dynamics error, cable-transmission error, and the one-sample command-hold error are additionally assumed bounded.

For $\boldsymbol e_R=\boldsymbol\theta-\boldsymbol\theta_{Ref}$ and $\boldsymbol e_v=\dot{\boldsymbol\theta}-\dot{\boldsymbol\theta}_{Ref}$, adding the human and robot dynamics and applying Eq.~(\ref{eq:robot_torque}) gives
\begin{equation}
\mathbf M\dot{\boldsymbol e}_v+\mathbf C\boldsymbol e_v
+\mathbf B_R\boldsymbol e_v+\mathbf K_R\boldsymbol e_R
=\boldsymbol d_{eff},
\label{eq:tracking_error_dynamics}
\end{equation}
where
\begin{equation}
\begin{aligned}
\boldsymbol d_{eff}={}&\boldsymbol\tau_{vol}
+\boldsymbol\tau_{Assi}
-\mathbf M\ddot{\boldsymbol\theta}_{Ref}
-\mathbf C\dot{\boldsymbol\theta}_{Ref}-\mathbf G\\
&+\boldsymbol d_{model}+\boldsymbol d_{hold},
\qquad \|\boldsymbol d_{eff}\|\le\bar d_{eff}.
\end{aligned}
\label{eq:effective_disturbance}
\end{equation}
The bound follows from the capability and assistance limits, the boundedness of $\boldsymbol\tau_{vol}$ argued above, the compact workspace, the reference-acceleration constraint, and the bounded-error assumptions. To establish boundedness of both position and velocity errors, rather than velocity alone, consider the strict Lyapunov function
\begin{equation}
V_{\varepsilon}=
\frac{1}{2}\boldsymbol e_v^T\mathbf M\boldsymbol e_v
+\frac{1}{2}\boldsymbol e_R^T\mathbf K_R\boldsymbol e_R
+\varepsilon\boldsymbol e_R^T\mathbf M\boldsymbol e_v .
\label{eq:strict_lyapunov}
\end{equation}
For sufficiently small $\varepsilon>0$, there exist positive constants $c_l,c_u,c_1,c_2$ such that, with $\boldsymbol z_e=[\boldsymbol e_R^T,\boldsymbol e_v^T]^T$,
\begin{equation}
c_l\|\boldsymbol z_e\|^2\le V_{\varepsilon}
\le c_u\|\boldsymbol z_e\|^2,
\qquad
\dot V_{\varepsilon}\le-c_1\|\boldsymbol z_e\|^2
+c_2\bar d_{eff}\|\boldsymbol z_e\|.
\label{eq:practical_stability_bound}
\end{equation}
Consequently, the tracking state is uniformly ultimately bounded, with
\[
\limsup_{t\rightarrow\infty}\|\boldsymbol z_e(t)\|
\le(c_2/c_1)\bar d_{eff}.
\]
For a fixed reference and zero effective disturbance, the same inequality gives asymptotic convergence.

\section{Experiments}
Two experiments were designed to evaluate the proposed method. Experiment~1 systematically varied movement intention and effective voluntary capability as separable factors, testing whether the dual-estimation architecture and attribution gate can distinguish these two sources of perturbation. Experiment~2 embedded the controller in a functional cup-transport task with a pre--post fatigue protocol, assessing its practical feasibility in a controlled ADL-like setting.

\subsection{Experimental Setup}
\subsubsection{Participants and Equipment}
\begin{figure}[H]
    \centering
    \includegraphics[width=0.82\textwidth]{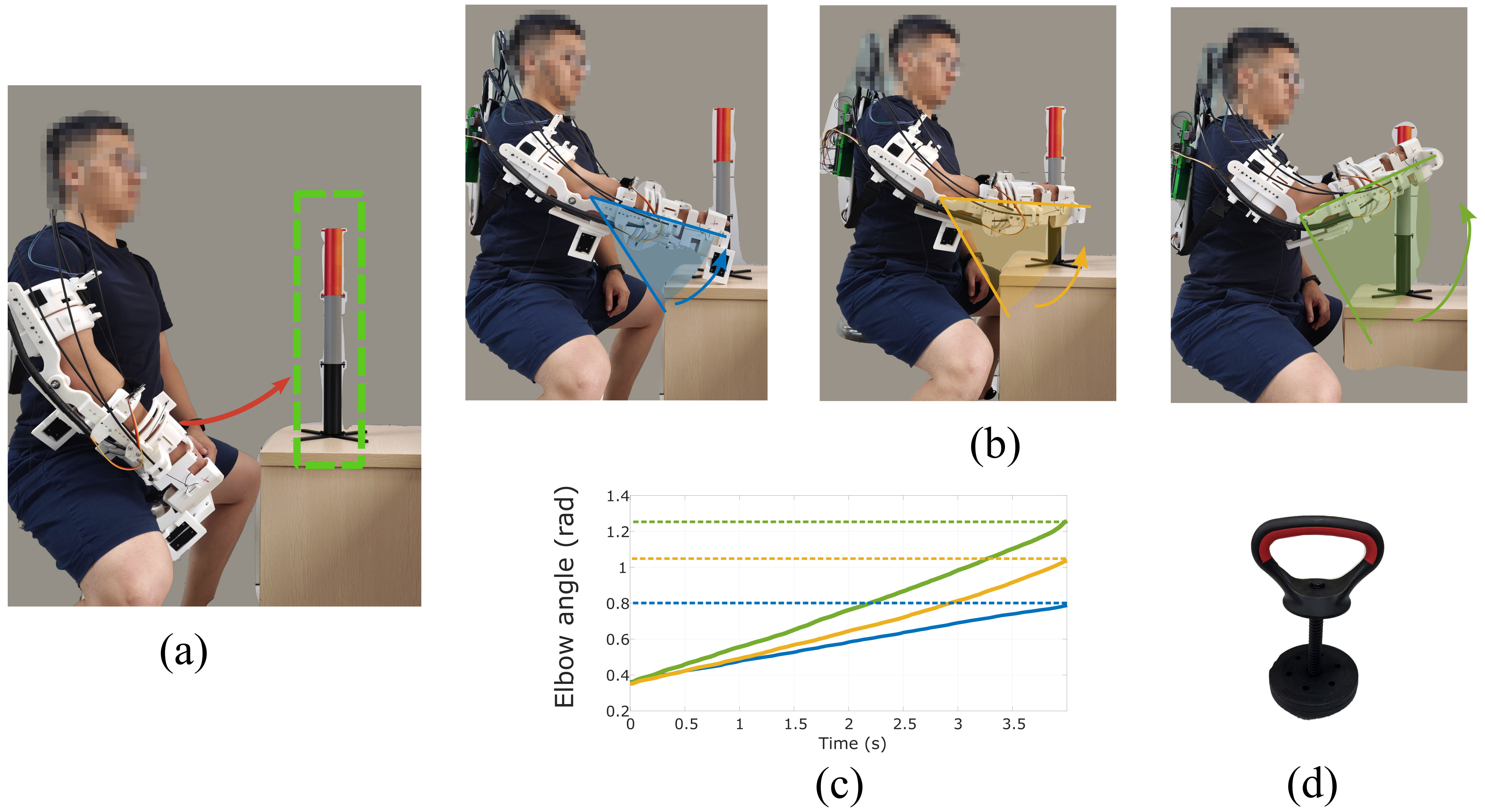}
    \caption{Experimental setup. (a) Participant wearing the upper-limb robot. (b) Three target heights used to vary movement intention. (c) Representative elbow trajectories. (d) Wrist-mounted loads used to vary task demand.}
    \label{fig:experiment_setup}
\end{figure}

\begin{figure}[H]
    \centering
    \includegraphics[width=0.82\textwidth]{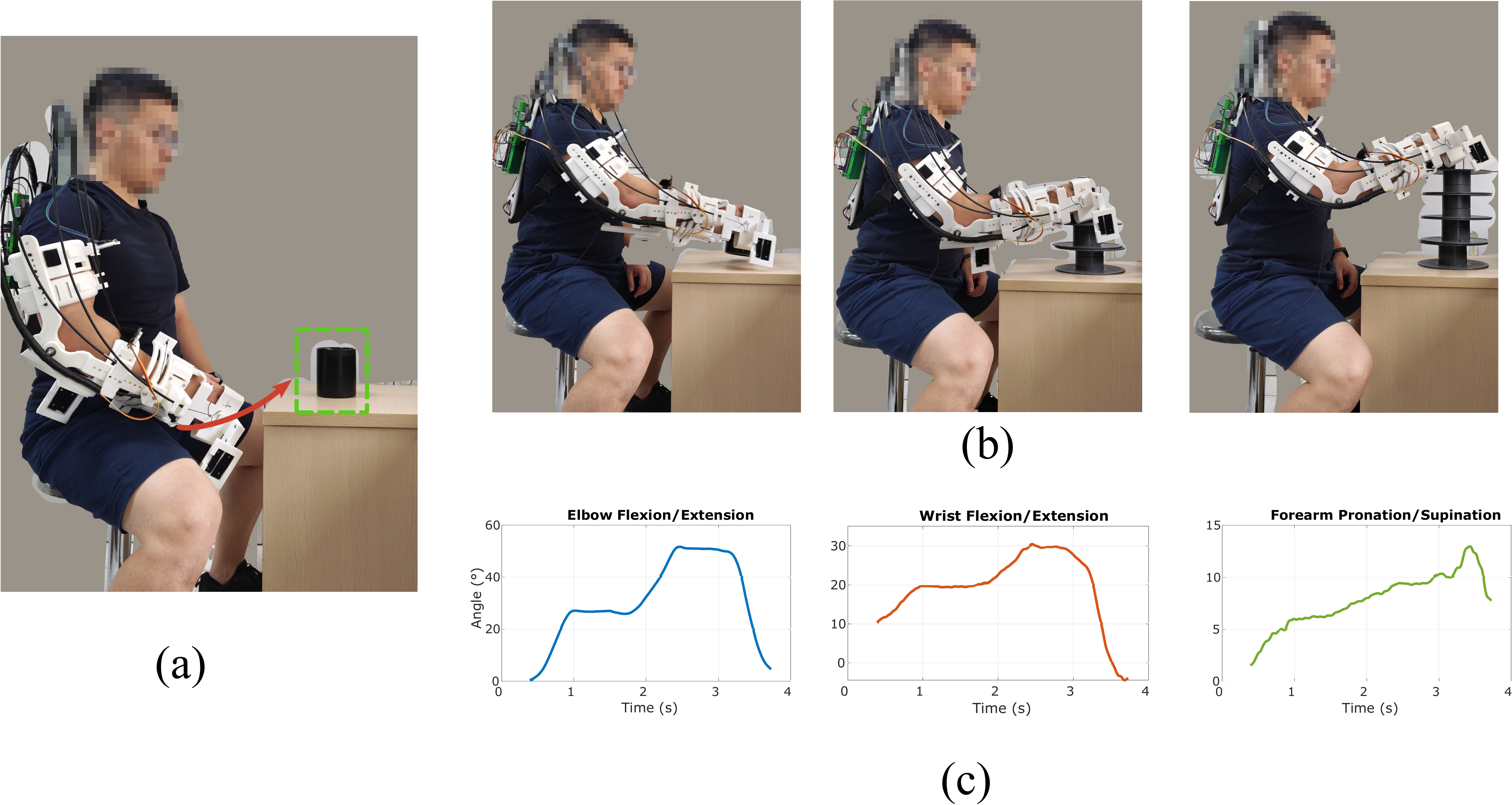}
    \caption{Experiment~2 functional cup-transport task. (a) A seated participant wearing the upper-limb robot reaches from the initial posture toward the cup. (b) Cup placement at the Low, Middle, and High target heights. (c) Representative joint-angle profiles for elbow flexion/extension, wrist flexion/extension, and forearm pronation/supination during one transport trial.}
    \label{fig:experiment2_setup}
\end{figure}

The robot was configured for two seated tasks. Experiment~1 used repeated reaching movements across target-height and wrist-load conditions (Fig.~\ref{fig:experiment_setup}), whereas Experiment~2 used a functional cup-transport task requiring coordinated elbow, wrist, and forearm motion (Fig.~\ref{fig:experiment2_setup}). Adaptive assistance was applied at the elbow in both experiments.

Nine healthy young participants (age: 24.3 $\pm$ 4.6 years, height: 171.4 $\pm$ 5.3 cm, weight: 62.1 $\pm$ 21.0 kg; two females) were recruited. All nine participants completed both experiments. All participants provided written informed consent under a protocol approved by the Institutional Review Board of Southern University of Science and Technology (No. 20250066). Participants were screened for neurological or musculoskeletal disorders, were right-handed, and wore the robot on their dominant arm.

\subsubsection{Individual Calibration}
\label{sec:participant_calibration}
Before formal testing, each participant completed three short calibration stages: resting sEMG recording, three EMG-only maximum voluntary contraction (MVC) attempts, and three comfortable unassisted reaches to the middle target. sEMG was recorded from the biceps brachii and triceps brachii using bipolar sensors (Sichiray, Wuxi, China; sampling rate: 1000~Hz) with Ag/AgCl electrodes. The calibration identified the participant-specific EMG baseline and activation scale, nominal middle-target trajectory, initial effective voluntary-torque capability, and associated uncertainty. Neither target-height information nor load data were used in calibration.

Elbow dynamic parameters were estimated from participant anthropometrics using standard ratios~\cite{ZYang2023Task}:
\begin{equation}
 m_f \approx 0.016\,m_{H}, \qquad l_f \approx 0.682\,L_{fa}
\end{equation}
where $m_H$ is body mass and $L_{fa}$ is forearm length. These parameters and the calibration-derived quantities initialized the participant-specific model. The assistance-onset factor was fixed at $\kappa=0.5$ for all reported experiments.

\subsection{Experiment 1: Controlled Variation of Intention and Capability}
To evaluate the separate response to intention and capability changes, target height (Low, Middle, High) was varied to induce intended trajectory deviations, while unmodeled wrist loads (Free, Light, Heavy) were applied to alter the effective voluntary capability--demand balance. All conditions shared a common nominal trajectory derived from the Middle target; the controller had to infer any intended deviation online. Target changes occurred only within a fixed load, and load changes occurred only at block boundaries with unchanged target, ensuring that the two perturbation types were temporally decoupled.

Three controllers were compared within subjects: (1) error-based AAN (baseline without estimation), (2) ungated dual-estimation MPC (base architecture without credit assignment), and (3) attribution-gated dual-estimation MPC (the proposed method). This progressive comparison isolated the contributions of dual estimation and of the attribution gate. Each controller--load block consisted of nine randomized trials (3 targets $\times$ 3 repetitions), and the capability estimate was reset to the calibrated prior at each block boundary. The same target sequence was reused across controllers, yielding $3 \times 3 \times 3 \times 3 = 81$ trials per participant.

\subsection{Experiment 2: Functional Task Under Induced Fatigue}
The gated controller was further evaluated in a cup-transport task (Fig.~\ref{fig:experiment2_setup}) that combined elbow, wrist, and forearm motion. A functional reference was built from three phase-aligned demonstrations. Nine trials (3 heights $\times$ 3 repetitions) were performed before and after a fatigue-inducing weight-lifting procedure, with the controller unchanged. This design tested whether the estimator could track a fatigue-induced decline in effective capability and increase assistance accordingly, without explicit fatigue modeling. 

In this experiment, elbow flexion/extension, wrist flexion/extension, and forearm pronation/supination were assisted, while only the elbow joint assistance was adjusted by the proposed method. The other two joints were assisted with a fixed-gain impedance controller to maintain the functional task.

\subsection{Statistical Analysis}
Results are reported as mean $\pm$ sample standard deviation across the nine participants after within-participant averaging. Experiment~1 was analyzed using a three-way repeated-measures ANOVA with controller, load, and target height as within-participant factors. Experiment~2 was analyzed using a two-way repeated-measures ANOVA with fatigue state and target height as within-participant factors. Omnibus effects were evaluated first; significant effects were followed by paired comparisons with Holm correction. Statistical significance was set at $p<0.05$.

\section{Experimental Results}
\label{sec:results}

\subsection{Experiment 1: Dual Estimation and Credit Assignment}
The primary goal of Experiment~1 was to test whether the dual-estimation framework can distinguish intended trajectory changes from capability-related changes, and whether the attribution gate improves this separation. Three findings jointly address this goal.

First, the attribution-gated MPC achieved the lowest tracking RMSE in every load--height condition (Table~\ref{tab:exp1_results}), providing descriptive evidence for the benefit of jointly estimating and separating the two measurands. Its condition-level RMSE averaged 0.017~rad, compared with 0.041~rad for the independent-estimation variant and 0.127~rad for error-based AAN. Across the nine conditions, the respective RMSE ranges were 0.011--0.024, 0.029--0.057, and 0.067--0.183~rad. The repeated-measures analysis confirmed significant effects of controller ($p<0.001$) and controller--height interaction ($p=0.011$) on RMSE; both dual-estimation controllers significantly outperformed error-based AAN after Holm correction ($p \le 0.004$).

\begin{figure}[H]
    \centering
    \includegraphics[
        width=0.96\textwidth,
        height=0.68\textheight,
        keepaspectratio
    ]{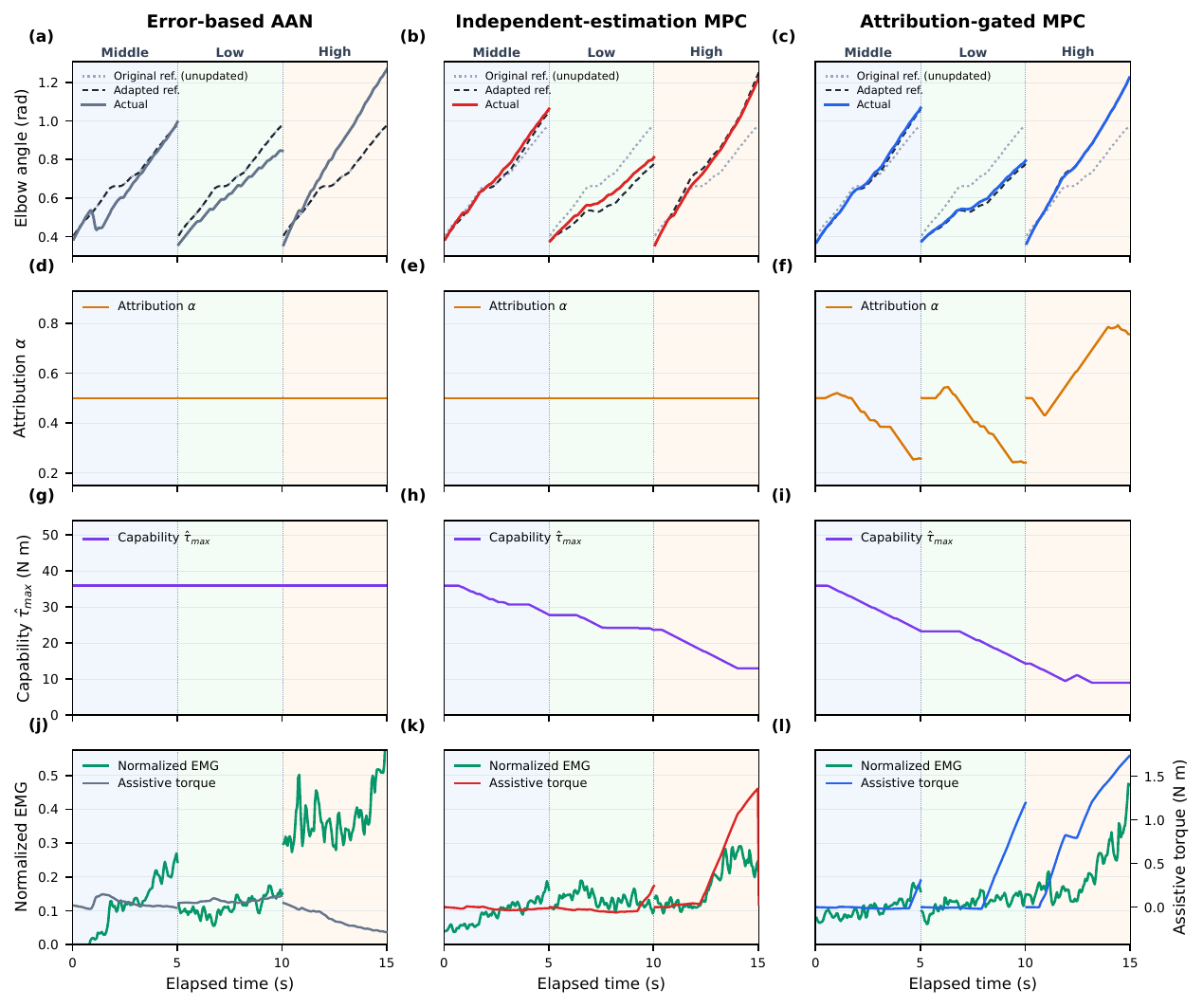}
    \caption{Continuous Experiment~1 responses from a representative recording, used to illustrate the controller behavior underlying the aggregate comparison. Signals are arranged by controller (columns) and signal group (rows). For each controller, consecutive raw trials in the Heavy-load block covering target heights Middle, Low, and High once each, retain that acquisition order, and were not selected based on response appearance. The top row overlays the measured elbow trajectory and the reference used by the controller; the repeated dotted trace is the same original, unupdated Middle-target nominal trajectory for all three target trials, whereas the dashed trace shows the resulting online-adapted reference. The next two rows show, on separate scales, logged attribution $\alpha$ and the capability estimate $\hat{\tau}_{max}$ in N$\cdot$m. The bottom row overlays normalized EMG (left axis) and commanded motor assistive torque (right axis, shown only on the rightmost panel). }
    
    \label{fig:result_controller_trials}
\end{figure}

\begin{table}[p]
\centering
\begin{minipage}{0.94\linewidth}
\caption{Experiment~1 outcomes.}
\label{tab:exp1_results}
\centering
\renewcommand{\arraystretch}{1.02}
\setlength{\tabcolsep}{3pt}
\resizebox{\linewidth}{!}{%
\begin{tabular}{lllcccccc}
\toprule
\textbf{Load} & \textbf{Height} & \textbf{Controller} &
\textbf{Actual endpoint (rad)} &
\textbf{Reference endpoint (rad)} &
\textbf{Tracking RMSE (rad)} &
\textbf{Mean attribution $\alpha$} &
\textbf{Normalized EMG} &
\textbf{Motor assistive torque (N\,m)} \\
\midrule
% BEGIN AUTO-GENERATED EXPERIMENT 1 LOAD-CONTROLLER ROWS
\multirow{9}{*}{Free (0 kg)} & \multirow{3}{*}{Low} & Error-based AAN & 0.783 $\pm$ 0.037 & 0.980 $\pm$ 0.030 & 0.170 $\pm$ 0.028 & 0.500 $\pm$ 0.000 & 0.009 $\pm$ 0.004 & 0.151 $\pm$ 0.026 \\
 &  & Independent MPC & 0.730 $\pm$ 0.028 & 0.690 $\pm$ 0.035 & 0.037 $\pm$ 0.013 & 0.500 $\pm$ 0.000 & 0.015 $\pm$ 0.011 & 0.007 $\pm$ 0.022 \\
 &  & Attribution-gated MPC & 0.702 $\pm$ 0.034 & 0.686 $\pm$ 0.033 & 0.017 $\pm$ 0.014 & 0.573 $\pm$ 0.029 & 0.026 $\pm$ 0.028 & 0.004 $\pm$ 0.011 \\
\cmidrule(lr){2-9}
 & \multirow{3}{*}{Middle} & Error-based AAN & 1.018 $\pm$ 0.058 & 0.980 $\pm$ 0.030 & 0.067 $\pm$ 0.037 & 0.500 $\pm$ 0.000 & 0.015 $\pm$ 0.007 & 0.038 $\pm$ 0.047 \\
 &  & Independent MPC & 0.986 $\pm$ 0.025 & 0.972 $\pm$ 0.026 & 0.034 $\pm$ 0.018 & 0.500 $\pm$ 0.000 & 0.018 $\pm$ 0.012 & 0.017 $\pm$ 0.023 \\
 &  & Attribution-gated MPC & 1.002 $\pm$ 0.043 & 1.000 $\pm$ 0.048 & 0.013 $\pm$ 0.016 & 0.489 $\pm$ 0.034 & 0.026 $\pm$ 0.023 & 0.005 $\pm$ 0.010 \\
\cmidrule(lr){2-9}
 & \multirow{3}{*}{High} & Error-based AAN & 1.202 $\pm$ 0.128 & 0.980 $\pm$ 0.030 & 0.128 $\pm$ 0.017 & 0.500 $\pm$ 0.000 & 0.062 $\pm$ 0.014 & -0.093 $\pm$ 0.032 \\
 &  & Independent MPC & 1.223 $\pm$ 0.052 & 1.294 $\pm$ 0.111 & 0.057 $\pm$ 0.043 & 0.500 $\pm$ 0.000 & 0.027 $\pm$ 0.013 & 0.048 $\pm$ 0.040 \\
 &  & Attribution-gated MPC & 1.284 $\pm$ 0.088 & 1.312 $\pm$ 0.118 & 0.024 $\pm$ 0.018 & 0.617 $\pm$ 0.066 & 0.033 $\pm$ 0.016 & 0.020 $\pm$ 0.018 \\
\midrule
\multirow{9}{*}{Light} & \multirow{3}{*}{Low} & Error-based AAN & 0.774 $\pm$ 0.077 & 0.980 $\pm$ 0.030 & 0.181 $\pm$ 0.060 & 0.500 $\pm$ 0.000 & 0.059 $\pm$ 0.019 & 0.162 $\pm$ 0.055 \\
 &  & Independent MPC & 0.724 $\pm$ 0.021 & 0.686 $\pm$ 0.025 & 0.031 $\pm$ 0.010 & 0.500 $\pm$ 0.000 & 0.078 $\pm$ 0.015 & 0.281 $\pm$ 0.223 \\
 &  & Attribution-gated MPC & 0.713 $\pm$ 0.045 & 0.702 $\pm$ 0.044 & 0.011 $\pm$ 0.010 & 0.549 $\pm$ 0.068 & 0.066 $\pm$ 0.028 & 0.375 $\pm$ 0.322 \\
\cmidrule(lr){2-9}
 & \multirow{3}{*}{Middle} & Error-based AAN & 0.999 $\pm$ 0.062 & 0.980 $\pm$ 0.030 & 0.093 $\pm$ 0.053 & 0.500 $\pm$ 0.000 & 0.097 $\pm$ 0.054 & 0.053 $\pm$ 0.082 \\
 &  & Independent MPC & 0.980 $\pm$ 0.036 & 0.973 $\pm$ 0.028 & 0.029 $\pm$ 0.012 & 0.500 $\pm$ 0.000 & 0.095 $\pm$ 0.030 & 0.250 $\pm$ 0.193 \\
 &  & Attribution-gated MPC & 0.991 $\pm$ 0.037 & 0.989 $\pm$ 0.037 & 0.012 $\pm$ 0.014 & 0.377 $\pm$ 0.045 & 0.082 $\pm$ 0.042 & 0.254 $\pm$ 0.213 \\
\cmidrule(lr){2-9}
 & \multirow{3}{*}{High} & Error-based AAN & 1.208 $\pm$ 0.054 & 0.980 $\pm$ 0.030 & 0.125 $\pm$ 0.043 & 0.500 $\pm$ 0.000 & 0.177 $\pm$ 0.050 & -0.081 $\pm$ 0.071 \\
 &  & Independent MPC & 1.237 $\pm$ 0.033 & 1.306 $\pm$ 0.092 & 0.053 $\pm$ 0.038 & 0.500 $\pm$ 0.000 & 0.110 $\pm$ 0.033 & 0.395 $\pm$ 0.285 \\
 &  & Attribution-gated MPC & 1.298 $\pm$ 0.089 & 1.330 $\pm$ 0.118 & 0.023 $\pm$ 0.018 & 0.745 $\pm$ 0.106 & 0.108 $\pm$ 0.042 & 0.443 $\pm$ 0.396 \\
\midrule
\multirow{9}{*}{Heavy} & \multirow{3}{*}{Low} & Error-based AAN & 0.762 $\pm$ 0.070 & 0.980 $\pm$ 0.030 & 0.183 $\pm$ 0.067 & 0.500 $\pm$ 0.000 & 0.244 $\pm$ 0.107 & 0.169 $\pm$ 0.061 \\
 &  & Independent MPC & 0.749 $\pm$ 0.036 & 0.700 $\pm$ 0.043 & 0.040 $\pm$ 0.026 & 0.500 $\pm$ 0.000 & 0.245 $\pm$ 0.104 & 0.428 $\pm$ 0.365 \\
 &  & Attribution-gated MPC & 0.722 $\pm$ 0.052 & 0.708 $\pm$ 0.051 & 0.015 $\pm$ 0.023 & 0.435 $\pm$ 0.074 & 0.224 $\pm$ 0.098 & 0.491 $\pm$ 0.432 \\
\cmidrule(lr){2-9}
 & \multirow{3}{*}{Middle} & Error-based AAN & 0.989 $\pm$ 0.083 & 0.980 $\pm$ 0.030 & 0.080 $\pm$ 0.023 & 0.500 $\pm$ 0.000 & 0.267 $\pm$ 0.125 & 0.043 $\pm$ 0.036 \\
 &  & Independent MPC & 0.989 $\pm$ 0.035 & 0.986 $\pm$ 0.038 & 0.032 $\pm$ 0.030 & 0.500 $\pm$ 0.000 & 0.258 $\pm$ 0.130 & 0.389 $\pm$ 0.316 \\
 &  & Attribution-gated MPC & 0.991 $\pm$ 0.041 & 0.991 $\pm$ 0.038 & 0.013 $\pm$ 0.026 & 0.402 $\pm$ 0.068 & 0.256 $\pm$ 0.137 & 0.379 $\pm$ 0.319 \\
\cmidrule(lr){2-9}
 & \multirow{3}{*}{High} & Error-based AAN & 1.202 $\pm$ 0.140 & 0.980 $\pm$ 0.030 & 0.120 $\pm$ 0.050 & 0.500 $\pm$ 0.000 & 0.576 $\pm$ 0.172 & -0.083 $\pm$ 0.055 \\
 &  & Independent MPC & 1.234 $\pm$ 0.048 & 1.315 $\pm$ 0.120 & 0.056 $\pm$ 0.044 & 0.500 $\pm$ 0.000 & 0.313 $\pm$ 0.139 & 0.621 $\pm$ 0.494 \\
 &  & Attribution-gated MPC & 1.273 $\pm$ 0.077 & 1.298 $\pm$ 0.102 & 0.021 $\pm$ 0.025 & 0.606 $\pm$ 0.166 & 0.255 $\pm$ 0.058 & 0.689 $\pm$ 0.577 \\
% END AUTO-GENERATED EXPERIMENT 1 LOAD-CONTROLLER ROWS
\bottomrule
\end{tabular}}
\vspace{2pt}

\begin{minipage}{\linewidth}
\footnotesize
\textit{Note:} Values are reported as mean $\pm$ sample standard deviation across nine participants, after first averaging repetitions within each participant at each load--height--controller condition. Low, Middle, and High target heights are reported separately. Actual and reference endpoints are the respective mean angles over normalized phase 0.90--1.00. Tracking RMSE is calculated from paired finite actual/reference phase samples; attribution, normalized EMG, and signed motor assistance are phase-sample means. Normalized EMG is reported from the biceps brachii as the primary elbow flexor. 
\end{minipage}
\end{minipage}
\end{table}

The representative Heavy-load trials in Fig.~\ref{fig:result_controller_trials} illustrate the mechanism behind this ordering. Because error-based AAN does not deform the reference, its reference remains superimposed on the repeated nominal trajectory and substantial target-dependent mismatch remains in the Low and High trials. Both dual-estimation controllers instead deform the reference downward for Low and upward for High. The gated controller's measured trajectory remains closely aligned with these adapted references, consistent with its lower aggregate RMSE in Table~\ref{tab:exp1_results}.

Second, the attribution state $\alpha$ varied systematically with load and height (Table~\ref{tab:exp1_results}), confirming that the gate dynamically assigned credit between the two update channels rather than remaining at a fixed neutral value. Under the nominal Free--Middle condition, mean $\alpha$ stayed near 0.489. Adding load at the same Middle target decreased $\alpha$ (0.377 Light, 0.402 Heavy), shifting weight toward the capability channel. Conversely, changing target under Free load shifted $\alpha$ upward (0.573 Low, 0.617 High), increasing trajectory-channel weight. The continuous traces in Fig.~\ref{fig:result_controller_trials} show that these state adjustments occurred within trials and across successive trials, consistent with the intended credit-assignment mechanism.

Third, the load manipulation produced the expected rise in normalized EMG of the biceps brachii, the primary elbow flexor (e.g., for the gated controller: Free 0.028, Light 0.085, Heavy 0.245), yet the height-averaged tracking RMSE of the gated controller remained low and similar across loads (0.018, 0.015, and 0.016~rad, respectively). This dissociation between increased effort and stable tracking indicates that the capability estimate and resulting assistance compensated for the added demand. The assistance torque itself showed a significant controller--height interaction ($p<0.001$), reflecting condition-specific modulation rather than a uniform load-driven increase. The actual-versus-reference endpoint differences in Table~\ref{tab:exp1_results}, together with the repeated unupdated nominal traces in Fig.~\ref{fig:result_controller_trials}, further show that the online reference was deformed toward the cued target height even though the controller received no target-height input.

\subsection{Experiment 2: Functional Task Under Fatigue}
\begin{figure}[H]
    \centering
    \includegraphics[width=0.72\textwidth]{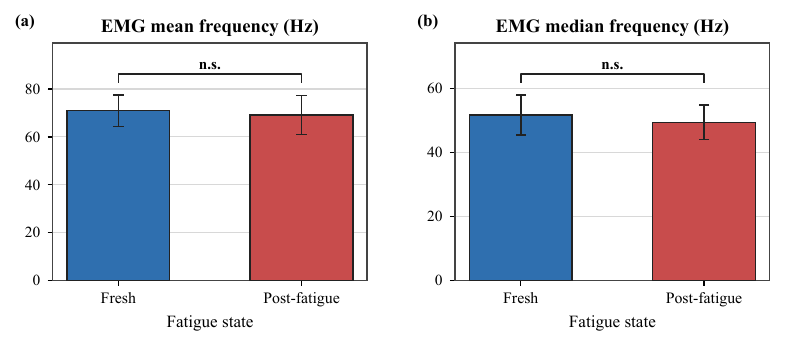}
    \caption{Experiment~2 EMG (a) mean and (b) median frequency before and after fatigue, averaged across targets and repetitions within each participant.}
    \label{fig:result_fatigue_emg}
\end{figure}

Experiment~2 tested whether the gated controller can track a fatigue-induced decline in effective capability and respond with increased assistance during a functional cup-transport task. The results showed consistent directional changes: after fatigue, the terminal effective capability $\hat{\tau}_{max}$ decreased at all three placement heights (by 8.005--10.783~N\,m across heights), while trial-mean assistance increased (by 0.249--0.300~N\,m) and tracking RMSE rose (by 0.007--0.025~rad) (Table~\ref{tab:exp2_results}). The mean and median sEMG frequencies decreased (70.94 to 69.20~Hz and 51.74 to 49.39~Hz, respectively; Fig.~\ref{fig:result_fatigue_emg}), independently corroborating the presence of fatigue, although these EMG shifts did not reach significance.

\begin{figure}[H]
    \centering
    \includegraphics[width=0.96\textwidth]{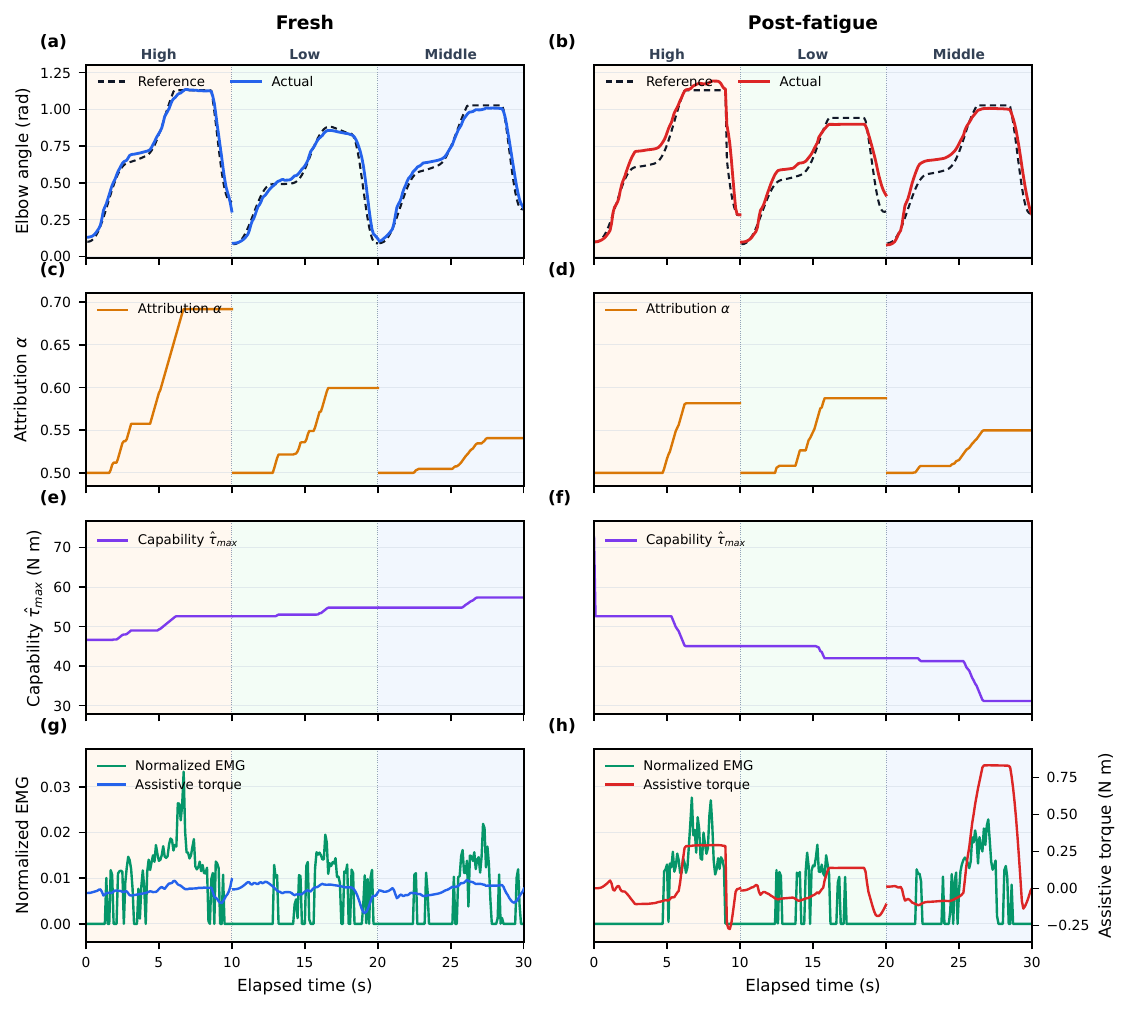}
    \caption{Representative Experiment~2 responses for one participant before and after fatigue. Each column contains three consecutive target-height trials in acquisition order; rows show elbow tracking, $\alpha$, $\hat{\tau}_{max}$, and normalized EMG/assistance. }
    \label{fig:result_fatigue_trials}
\end{figure}

Only the increase in RMSE was statistically significant ($p<0.001$), while the capability and assistance changes showed consistent trends across participants without reaching significance ($p=0.331$ and $p=0.488$, respectively). The continuous trial data in Fig.~\ref{fig:result_fatigue_trials} illustrate the coordinated controller response: a reduction in $\hat{\tau}_{max}$ and an increase in assistance accompanied greater tracking deviation after fatigue.

\begin{table}[H]
\caption{Experiment~2 outcomes by target height before and after fatigue (mean $\pm$ sample standard deviation; $n=9$).}
\label{tab:exp2_results}
\centering
\renewcommand{\arraystretch}{1.06}
\setlength{\tabcolsep}{2.2pt}
\resizebox{\columnwidth}{!}{%
\begin{tabular}{@{}llccc@{}}
\toprule
\textbf{Target} & \textbf{State} &
\textbf{$\hat{\tau}_{max}$ (N\,m)} &
\textbf{Mean assist (N\,m)} &
\textbf{RMSE (rad)} \\
\midrule
% BEGIN AUTO-GENERATED EXPERIMENT 2 FATIGUE ROWS
\multirow{2}{*}{Low} & Fresh & 47.240 $\pm$ 11.950 & -0.005 $\pm$ 0.004 & 0.041 $\pm$ 0.022 \\
 & Post-fatigue & 36.457 $\pm$ 20.231 & 0.244 $\pm$ 0.380 & 0.062 $\pm$ 0.012 \\
\midrule
\multirow{2}{*}{Middle} & Fresh & 46.734 $\pm$ 12.298 & 0.009 $\pm$ 0.033 & 0.053 $\pm$ 0.023 \\
 & Post-fatigue & 36.260 $\pm$ 16.491 & 0.282 $\pm$ 0.325 & 0.060 $\pm$ 0.011 \\
\midrule
\multirow{2}{*}{High} & Fresh & 45.268 $\pm$ 9.308 & 0.069 $\pm$ 0.077 & 0.056 $\pm$ 0.016 \\
 & Post-fatigue & 37.263 $\pm$ 20.503 & 0.369 $\pm$ 0.530 & 0.081 $\pm$ 0.015 \\
% END AUTO-GENERATED EXPERIMENT 2 FATIGUE ROWS
\bottomrule
\end{tabular}}
\end{table}

\section{Discussion}

Recent studies have advanced individual components of this problem but do not resolve the ambiguity at the closed-loop control level. Multitask networks recognize motion intention and fatigue simultaneously from sEMG~\cite{Tu2025TIM}, and separate work detects fatigue by fusing time- and frequency-domain features~\cite{Wu2025TIMFatigue}, but these outputs are discrete labels rather than continuous control-relevant measurands. Simultaneous intention--stiffness estimation has been demonstrated for physical human--robot interaction~\cite{Wu2025}, yet stiffness characterizes mechanical impedance rather than voluntary torque capability. Reference-adaptation methods~\cite{Han2025,Franceschi2025} and capability-centered AAN schemes~\cite{Shah2025Shoulder,Zarrin2024VariableAdmittance} therefore assign deviations to one prespecified channel. The unresolved issue is structural: when intention and capability share kinematic, interaction, and sEMG observations, parallel estimators cannot determine which latent cause should receive credit for a new residual. Joint estimation is needed to expose both hypotheses to the same evidence, and attribution is needed to prevent the same evidence from driving both estimates simultaneously.

\subsection{Simultaneous Estimation of Intention and Capability}
A central finding is that intended trajectory deformation and task-dependent effective voluntary capability can be updated online as distinct, control-relevant quantities with condition-consistent responses. This goes beyond recent simultaneous recognition results~\cite{Wu2025, Tu2025TIM}: it produces two continuous measurands and explicitly models their competing explanations of the observations. In Experiment~1, the reference endpoints of the gated controller remained correctly ordered (Low, Middle, High) under all three load conditions, despite receiving no target-height information and using the same Middle-target nominal trajectory. The updated first row of Fig.~\ref{fig:result_controller_trials} makes this distinction explicit: the dotted original trajectory repeats across the three targets, whereas the online reference deforms downward for Low and upward for High and remains close to the measured path. Thus, the target-specific reference changes cannot be attributed to different original trajectories. At the same time, $\hat{\tau}_{max}$ progressively declined during the Heavy-load block, showing that the capability channel continued to respond to the added demand while the reference followed the cued target transitions.

The capability estimator's response to unmodeled load also explains why tracking RMSE remained low across load-averaged values for the gated controller. Because the MHE continuously updates $\hat{\tau}_{max}$ from the EMG--dynamics residual, the MPC can anticipate the required assistance without relying on a growing tracking error. Recent work has shown that EMG-to-force mappings can be adapted across fatigue states~\cite{Pan2025TNSRE}, and that physiological and functional measurements capture complementary aspects of fatigue~\cite{Wu2025TIMFatigue,Pitzalis2025TNSRE}. However, estimating force or detecting fatigue alone cannot tell an AAN controller whether a task mismatch should increase assistance or deform the reference. The proposed paired measurands provide that missing decision context. This predictive compensation decouples assistance from raw kinematic deviation---a property that error-based AAN fundamentally lacks. The significant controller--height interaction in assistance torque further confirms that the commanded torque was modulated by task context rather than by a uniform, load-driven offset.

\subsection{Dynamic Credit Assignment Through the Attribution State}
The attribution gate $\alpha$ prevents cross-contamination between the two estimation channels. This role differs from feature sharing in multitask recognition~\cite{Tu2025TIM} and from cost arbitration in shared-control adaptation~\cite{Han2025,Franceschi2025}: the gate acts on the causal destination of a residual before it changes the reference or assistance demand. Under the nominal Free--Middle condition, $\alpha$ remained near neutral (0.489). Adding load at the same target decreased $\alpha$ (0.377 Light, 0.402 Heavy), shifting weight toward the capability channel, because the load increased the EMG--dynamics residual without producing a directionally consistent corrective torque. Changing target under Free load shifted $\alpha$ upward (0.573 Low, 0.617 High), as persistent corrective interaction provided clear trajectory-change evidence. These shifts were continuous, allowing the gate to express varying confidence and combine evidence when both intention and capability changed simultaneously.

The temporal dynamics in Fig.~\ref{fig:result_controller_trials} illustrate this arbitration against an unchanged nominal-trajectory baseline. During the displayed Middle and Low Heavy-load trials, decreases in $\alpha$ accompanied downward revisions of $\hat{\tau}_{max}$, placing more weight on capability evidence. In the subsequent High trial, $\alpha$ rose strongly while the adapted reference moved upward away from the repeated nominal trace; capability and assistance continued to change as well. The response therefore reflects continuous assignment when trajectory and capability evidence coexist, rather than a sequence of isolated hard classifications. This temporal coordination is consistent with the gate operating as a supervisory scheduler that limits capability deficits from being absorbed as intended deformations and deliberate target changes from being interpreted solely as capability loss. The RMSE improvement of the gated over the independent-estimation variant, although not significant after correction, aligns with this mechanistic interpretation and warrants further investigation with larger sample sizes.

\subsection{Controlled Evidence and Functional Feasibility}
The two experiments provided complementary validation. Experiment~1 employed a full factorial manipulation of target height and load, creating a $3 \times 3$ grid in which intention and capability perturbations were applied independently and in combination. This allowed controller-state responses to be mapped directly onto known perturbation types, establishing strong internal validity.

Experiment~2 extended this validation to an ecologically more relevant setting involving multi-joint coordination and a natural movement sequence under fatigue. This is important because recent high-performing sEMG fatigue models have predominantly been validated under structured contractions~\cite{Tu2025TIM,Wu2025TIMFatigue}, and a recent review identifies context, sensing modality, and calibration as persistent obstacles to interpreting fatigue during functional performance~\cite{Pitzalis2025TNSRE}. Although the pre--post changes in $\hat{\tau}_{max}$ and assistance did not reach significance, their consistent direction across all participants and target heights, together with the significant RMSE increase and sEMG spectral compression, strongly suggests that the controller detected and responded to the fatigue-induced deficit. The continuous traces in Fig.~\ref{fig:result_fatigue_trials} confirm that the coordinated state adaptations observed in controlled reaching were preserved during functional movement. This transfer to an ADL-like task supports the practical feasibility of the proposed framework.

\subsection{Limitations and Future Work}
The study enrolled healthy participants and used external loads and acute fatigue to simulate capability changes. While this enabled controlled, repeatable perturbations, it does not reproduce the complex neuromuscular impairments or long-term adaptation of clinical populations. Moreover, no single biosignal fully characterizes central and peripheral fatigue~\cite{Pitzalis2025TNSRE}. The results therefore validate algorithmic function rather than therapeutic efficacy. Subsequent studies should involve individuals with upper-limb motor deficits to evaluate patient-specific adaptation.

Second, adaptive capability estimation and assistance were restricted to elbow flexion. The cup-transport task involved wrist and forearm motion, but capability changes in those degrees of freedom were not estimated or compensated. Extending the framework to multiple joints will require managing the increased dimensionality of the MHE--MPC problem and interactions between joint-level attribution states, which is an important direction for future work.

\section{Conclusion}
This work addresses a fundamental ambiguity in assist-as-needed control---that similar kinematic and interaction responses can arise from changes in movement intention or from reduced voluntary capability---by simultaneously estimating both quantities online and dynamically routing shared sensory evidence through an attribution gate. The key contribution lies in the credit-assignment mechanism that prevents intention- and capability-related signals from contaminating each other, enabling the controller to adapt the reference trajectory and adjust assistance appropriately without requiring hard event classification.

The attribution-gated controller achieved a condition-level mean tracking RMSE of 0.017~rad across nine load--height conditions, approximately 60\% lower than the independent-estimation variant and 87\% lower than error-based AAN. The attribution state $\alpha$ varied systematically with load and target height, while the continuous responses showed target-specific reference deformation away from a repeated unupdated nominal trajectory together with coordinated capability and assistance changes. A functional cup-transport task under induced fatigue provided further proof-of-concept evidence in a multi-joint ADL-like context: post-fatigue capability estimates decreased and assistance increased at all target heights, although these two changes were not statistically significant. Together, these findings support the proposed credit-assignment framework while motivating larger studies in clinical populations.

\section*{Ethics approval and informed consent}
The study protocol was approved by the Institutional Review Board of Southern University of Science and Technology (No. 20250066). All participants provided written informed consent before participation.

\section*{Funding}
This work was supported by the High Level Special Funds under Grant G03034K003 from Southern University of Science and Technology, Shenzhen, China, and by the Shenzhen ``Medical Engineering Integration'' Special Support Program Project under Grant F-2024-Z99-502365, Shenzhen, China.

\section*{Declaration of competing interest}
The authors declare that they have no known competing financial interests or personal relationships that could have appeared to influence the work reported in this paper.

\section*{Data availability}
The data that support the findings of this study are available from the corresponding author upon reasonable request.

% CRediT roles should be entered in the Editorial Manager submission form or
% added here after every author has confirmed the role allocation.

\end{document}